%% file: main.tex
\documentclass[letterpaper, 10 pt, conference]{ieeeconf}  %

\IEEEoverridecommandlockouts
\usepackage{cite}
\usepackage{amsmath,amssymb,amsfonts}
\usepackage{algorithmic}
\usepackage{graphicx}
\usepackage{textcomp}
\usepackage{xcolor}
\usepackage{balance}
\usepackage{comment}
\usepackage[ruled,vlined]{algorithm2e}
\usepackage{subcaption}
\usepackage{tensor}

\newcommand{\ie}{\emph{i.e.},}
\newcommand{\etal}{\emph{et~al.}}
\usepackage{hyperref}

\def\BibTeX{{\rm B\kern-.05em{\sc i\kern-.025em b}\kern-.08em
    T\kern-.1667em\lower.7ex\hbox{E}\kern-.125emX}}
\begin{document}

\title{\LARGE \bf Locus: LiDAR-based Place Recognition using Spatiotemporal Higher-Order Pooling
}

\author{Kavisha Vidanapathirana$^{1,2}$, Peyman Moghadam$^{1,2}$, Ben Harwood$^{1}$, Muming Zhao$^{1}$, \\ Sridha Sridharan$^{2}$, Clinton Fookes$^{2}$   
\thanks{
$^1$ Kavisha Vidanapathirana, Peyman Moghadam, Ben Harwood, Muming Zhao are with the Robotics and Autonomous Systems Group, DATA61, CSIRO, Brisbane, QLD 4069, Australia. 
E-mails: {\tt\footnotesize \emph{
firstname.lastname
}@data61.csiro.au}}
\thanks{
$^{2}$ Kavisha Vidanapathirana, Peyman Moghadam, Sridha Sridharan, Clinton Fookes are with the School of Electrical Engineering and Robotics, Queensland University of Technology (QUT), Brisbane, Australia.
E-mails: {\tt\footnotesize \emph\{kavisha.vidanapathirana, peyman.moghadam, s.sridharan, c.fookes\}@qut.edu.au}}
}

\maketitle

\input{chapters/abstract}

\input{chapters/introduction.tex}

\input{chapters/related_work.tex}

\input{chapters/method.tex}
\input{chapters/experiments.tex}
\input{chapters/results.tex}

\input{chapters/conclusion.tex}

\balance{}

\bibliographystyle{./bibliography/IEEEtran}
\bibliography{main.bbl}

\end{document}

%% file: chapters/abstract.tex
\begin{abstract}
Place Recognition enables the estimation of a globally consistent map and trajectory by providing non-local constraints in Simultaneous Localisation and Mapping (SLAM). This paper presents \textit{Locus}, a novel place recognition method using 3D LiDAR point clouds in large-scale environments. We propose a method for extracting and encoding topological and temporal information related to components in a scene and demonstrate how the inclusion of this auxiliary information in place description leads to more robust and discriminative scene representations. Second-order pooling along with a non-linear transform is used to aggregate these multi-level features to generate a fixed-length global descriptor, which is invariant to the permutation of input features. The proposed method outperforms state-of-the-art methods on the KITTI dataset. Furthermore, \textit{Locus} is demonstrated to be robust across several challenging situations such as occlusions and viewpoint changes in 3D LiDAR point clouds. The open-source implementation is available at: \href{https://github.com/csiro-robotics/locus}{https://github.com/csiro-robotics/locus} .

\end{abstract}

%% file: chapters/introduction.tex
\section{Introduction}
\label{sec:intro}

Place Recognition (PR) is an essential capability required by autonomous robots and driverless cars, which enables the recognition of previously visited places under changing viewpoints and environmental conditions. Place recognition is crucial for various applications such as loop closure detection for large-scale, global data association in Simultaneous Localization and Mapping (SLAM) or metric localization within known maps \cite{Park2018,park2019robust}. 

In this paper, we consider the problem of place recognition based on 3D LiDAR point clouds. The majority of the current state-of-the-art LiDAR-based place recognition methods extract representations of 3D point clouds based on either local or global descriptors. Local descriptors such as keypoint features \cite{Salti2014} often suffer from low repeatability under noise and minor changes in the environment. 
The demand for extracting discriminative and generalizable global descriptors from point clouds
has led to the development of several different techniques \cite{He2016,Uy2018,Kim2018}. 
However, the majority of global descriptors are not robust to viewpoint variations, occlusions and dynamic scenes \cite{Dube2017}. A potential reason for this fragility is that these methods are inherently not capable of capturing topological and temporal interactions between different components in the scene. Such auxiliary information is vital for the construction of robust and discriminative representations since a higher-level understanding is needed to distinguish between different scenes of similar structure.  

To this end, we propose a novel place recognition method (named \textit{Locus}) which effectively exploits multi-level features for global place description using 3D LiDAR data.  We define multiple levels of feature representations for each point cloud frame. A point cloud frame is defined as the points accumulated from a sweep of the LiDAR \cite{Geiger2013}. 
First, segment features are extracted to encode structural information 
of each point cloud frame. The segment-based representation leverages from the advantages of both local and global representations while not suffering from their individual drawbacks. Compared to local keypoint based representations, they are more descriptive and more robust to noise and changing environments. %
Next, spatial feature pooling encodes topological relationships between segments within a point cloud frame. Capturing topological relationships between components of a scene enables discrimination between scenes of similar composition but different arrangement. 

Finally, temporal feature pooling encodes co-occurrences of segments in the past $k_t$ point cloud frames providing robustness to sensor motion and dynamic objects.  
Once multi-level features are extracted, second-order pooling aggregates information of local features over a point cloud frame to form a holistic representation for place recognition.  Second-order pooling captures the multiplicative interactions between the multi-level features and outputs a fixed-length descriptor which is invariant to the number and permutation of the input features. Furthermore, the fixed dimension of the global descriptor enables maintaining the computational complexity.

\begin{figure*}[ht]
    \centering
    \includegraphics[width=17.5cm, height=5.5cm]{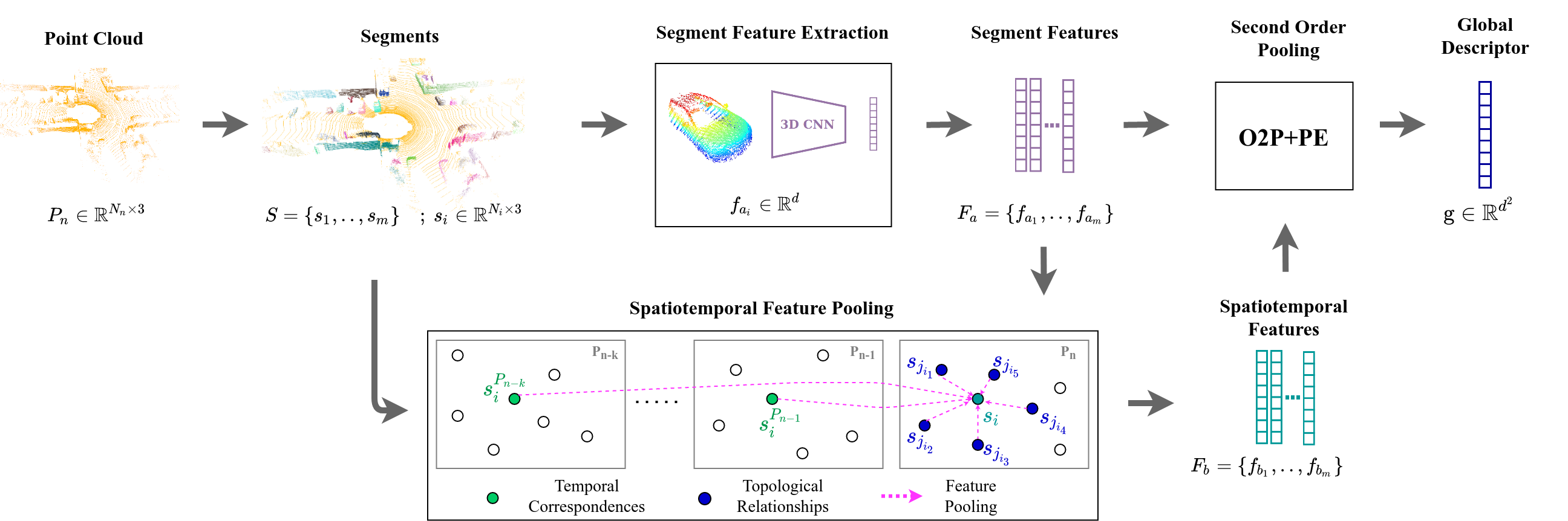}
    \caption{The overall framework of the proposed \textit{Locus} method. Segments extracted from a LiDAR point cloud frame are described using two complementary sets of features. One feature describes the structural appearance of a segment while the other encodes topological and temporal information related to a segment. The two sets of features are aggregated using second-order pooling (O2P) followed by a Power-Euclidean transform (PE) to obtain a global descriptor of the point cloud.}
    \label{fig:bl_pipeline}
\end{figure*}

Our main contributions are summarised as follows: 
\begin{itemize}
    \item We introduce multi-level features which encode structural appearance, topological relationships and temporal correspondences related to components in a scene.
    \item We formulate the generation of a global descriptor which encodes these multi-level features into a single viewpoint invariant representation using second-order pooling and demonstrate how these multi-level features contribute to place recognition performance. 
    
    \item Our proposed method (\textit{Locus}) outperforms the state-of-the-art place recognition methods on the KITTI dataset. 
    \item We quantitatively evaluate the robustness of our \textit{Locus} method on a variety of challenging scenarios such as viewpoint changes and occlusion. %
\end{itemize}

%% file: chapters/related_work.tex
\section{Related Work}
\label{sec:rel_work}

\subsection{Point cloud representation and descriptor generation}
To address the challenge of place recognition, the scene-level point cloud is often encoded in three different ways; a set of local descriptors, a single global descriptor or a set of object/segment descriptors. 
Local descriptor based methods first detect a set of keypoints in the point cloud and then form local descriptors by encoding information in each keypoint neighbourhood \cite{Salti2014,Guo2019}. Local descriptors and keypoint detection suffer from low repeatability in noisy point clouds and changing environments. 

Global descriptors aim to describe the entire point cloud by a single vector representation. M2DP~\cite{He2016} generates a global descriptor by projecting the points into multiple 2D planes and calculating density signatures. PointNetVLAD~\cite{Uy2018} extracts a global descriptor using an end-to-end process based on PointNet and NetVLAD \cite{Arandjelovic2018}. PointNetVLAD ignores the spatial distribution of features and hence lacks descriptive power.
LPD-Net \cite{lpdnet_new} addresses the limitation of PointNetVLAD by using adaptive local feature extraction along with a graph-based neighborhood aggregation module. Recently, DH3D \cite{Du2020} proposed a network which learns both the keypoint detection and local description and generates a global descriptor using NetVLAD. 
The aforementioned global descriptors do not
demonstrate rotational-invariance and often fail upon reverse revisits. 

The current state-of-the-art consists of several rotation-invariant global descriptors. ScanContext \cite{Kim2018} records the maximum height of points in a 2D grid-map of the scene and computes pairwise descriptor distances by comparing distances to all column shifted variants of the query descriptor to achieve rotational-invariance. Intensity ScanContext \cite{Wang2020a} extends ScanContext by including the intensity return of the LiDAR sensor. The ScanContext and its variants are not capable of capturing scene composition or topological relationships and rely on expensive distance calculation to ensure rotational-invariance. 

Recently, Kong~\etal{}~\cite{Kong2020} represented point clouds as a semantic graph and recognize places using a graph similarity network. Their network is capable of capturing topological and semantic information from the point cloud and also achieves rotational-invariance along with state-of-the-art performance. However, due to the use of semantic segmentation, their method suffers from the following two bottlenecks. First, it is dependant on the existence of pre-defined semantic classes in the test dataset (\ie{} domain gap problem).  Second, since each segment is only represented by its class label, the method is not capable of differentiating between two segments of the same class and thus loses valuable information related to intra-class variations.

Euclidean segment based representations \cite{Dube2017} are less affected by the aforementioned drawbacks while still being able to capture topological and semantic information of the components in a scene. 
SegMatch~\cite{Dube2017} and SegMap \cite{Dube2020} address the challenges of local and global descriptors by constructing a global map along the trajectory and performs recognition via segment-wise kNN retrieval on this map followed by a geometric consistency check. 
In this paper, we also leverage from segment descriptors for point cloud encoding but we avoid the construction of a global map and instead treat place recognition as a retrieval problem similar to recent LiDAR-based place recognition research \cite{Kim2018,Uy2018,Wang2020a,Kong2020}. 
In this sense, we avoid segment-wise kNN search and construct a global descriptor through the aggregation of  multi-level segment features. 

\subsection{Feature aggregation}

Bag-of-Words (BoW) \cite{cummins2008fab} and its higher-order variants (Vector of Locally Aggregated Descriptors (VLAD) \cite{Jegou2012}, NetVLAD \cite{Arandjelovic2018}, Fisher Vector (FV) \cite{Sanchez2013}) are popular methods in many retrieval tasks which construct global descriptors by aggregating local descriptors using zeroth, first or second order statistics respectively. 
These codebook-based aggregation methods require a preprocessing step to learn the codebook (a standard BoW, cluster centers in VLAD and Gaussian Mixture Models (GMM) in FV) and do not transfer well to unseen environments. %

Other higher-order aggregation methods such as Second Order Pooling (O2P) \cite{Carreira2012}
do not require such pre-processing steps and and hence enables to be used for online place recognition in unseen environments. They also tend to be more efficient in run-time since the description stage does not require finding the nearest codebook word for each feature. Additionally, O2P methods allow the aggregation of complementary features \cite{Lin2015} while maintaining a fixed-length descriptor without violating permutation invariance. 
In this paper, we propose to use second-order pooling to aggregate multiple levels of feature representations (structural, topological and temporal) into a fixed-length global descriptor for place recognition. 

\subsection{Incorporating temporal information}
In place recognition, temporal information has been used almost exclusively in the retrieval stage. In visual place recognition (VPR), SeqSLAM \cite{Milford2012}  incorporated sequential information by comparing feature similarity over time and demonstrated dramatic performance improvement when dealing with changing appearance. Lynen \etal{}~\cite{Lynen2015} modeled sequential retrieval as a probability density estimation in votes versus travel distance space. 
More recently, SeqLPD \cite{Liu2019b} extended LPD-Net with SeqSLAM for LiDAR-based place recognition. Our method differs from these sequential retrieval methods by incorporating temporal information in the place description stage instead. Recently, in VPR, Delta descriptors \cite{Garg2020} outlined how visual sequential-representation provides inherent robustness to variations in camera motion and environmental changes.

%% file: chapters/method.tex
\section{Method}
\label{sec:method}

This section describes our proposed method (named \textit{Locus}) for LiDAR-based place description including the segment based representation of the point cloud, the generation of the multi-level features and their second-order aggregation (see Fig.~\ref{fig:bl_pipeline}). 

\subsection{Segment-based representation}
\label{section:segmap}
Segments are defined as Euclidean clusters of points in the point cloud representation. 
Our segment extraction is performed similar to the  SegMatch implementation \cite{Dube2017} 
by first removing the ground plane and then extracting Euclidean clusters of points. The maximum Euclidean distance between two neighbouring points in the same segment is set to $ 0.2m $ and the segments contain a minimum of 100 and a maximum of 15000 points. Thus, each input point cloud $ P \in \mathbb{R}^{N \times 3} $ is represented by a set of segments
$ S = \{s_1, .., s_m\} $, where $ s_i \in \mathbb{R}^{N_i \times 3}$
, and the number of segments $ m $ varies depending on the environment and range of the point cloud.

For each segment, a compact deep feature which encodes the structural appearance of the segment is obtained using the \textit{SegMap-CNN} network proposed in \cite{Dube2020}. The network represents each segment in a voxel grid of 32x32x16 of which the voxel sizes (0.1m by default) are scaled in order to fit larger segments. The description network consists of three 3D convolutional layers plus max-pool layers followed by two fully connected layers. For a given set of $ m $ segments $ S $, it outputs a set of compact features
$ F_a = \{f_{a_1}, .., f_{a_m}\} $ (where $ f_{a_i} = f_a(s_i) \in \mathbb{R}^{d} , d = 64$)
which discriminates segments based on structural appearance. 

\subsection{Spatiotemporal Pooling}
Incorporating  topological and temporal information in place description has many advantages when dealing with changing environments and varying sensor motion. With the aim of encoding this information, 
we compute a complimentary set of features $ F_b \in \mathbb{R}^{m \times d} $ for each point cloud frame $ P_n $, in addition to the structural appearance features $ F_a \in \mathbb{R}^{m \times d} $ described in section \ref{section:segmap}. This is achieved via two stages of feature pooling: spatial and temporal. 

Feature pooling computes a new feature for a segment by taking the weighted average of features of all related segments. Encoding topological information is achieved via feature pooling based on spatial relationships within a point cloud frame. Temporal information is encoded via pooling features based on temporal correspondences across multiple point cloud frames. 
For a query segment $ s^{P_n}_i  $ from the current point cloud frame $ P_n $, we find topological relationships and temporal correspondences with all other segments $ s^{P_l}_j \in \mathbb{S}_{n,k_t} $ from the current frame and $ k_t $ frames into the immediate past, 
\begin{equation}
 \mathbb{S}_{n,k_t} = \{ S^{P_n}, .. , S^{P_{n-k_t}} \}, \quad S^{P_l} = \{ s^{P_l}_1, .. , s^{P_l}_{m_l} \},
\end{equation}
If a segment $ s^{P_l}_j  $ is spatially or temporally related to the query segment $  s^{P_n}_i  $, its structural appearance feature $ f_{a}(s^{P_l}_j) $ is included in the feature pooling of the query segment.

\subsubsection{\textbf{Spatial Feature Pooling}}
We define spatial relationships between segments through a directed graph $ \mathcal{G = (V,E)} $.
Vertices $ \mathcal{V} = \{1, .., m\} $ represent the segments $ S = \{s_1, .., s_m\} $  in the current point cloud frame and edges $ \mathcal{E} \subseteq \mathcal{V} \times \mathcal{V} $ represent which segments relate to which other segments.  $ \mathcal{G} $ is constructed as a kNN graph where each segment is connected to its $ k_s $ nearest neighbours ($ k_s = 5 $) and the distance between segments is calculated using the minimum translational distance (MTD) \cite{Cameron} between the convex hulls of the segments. We use the $ \mathtt{QuickHull} $ algorithm \cite{Barber1996} to compute convex hulls. Since the segment extraction process in Section \ref{section:segmap} guarantees that segments do not overlap, the MTD ($ \mathcal{D} $)  can be computed as follows,
\begin{equation}
\mathcal{D}(s_i,s_j) = \min \{ \|p_{i_x} - p_{j_y}\|: \forall \ p_{i_x} \in \hat{s_i}, p_{j_y} \in \hat{s_j} \}
\end{equation}
where $ \| p_{i_x} - p_{j_y} \| $ represents
Euclidean distance between points $ p_{i_x}, p_{j_y} \in \mathbb{R}^3  $ and
 $ \hat{s} = \mathtt{QuickHull}(s) $.
The spatial feature pooling for segment $ s_i $ is then carried out as,
\begin{equation} %
\Phi(s_i) = \sum_{j:(i,j) \in \mathcal{E} } \phi(i,j)f_{a}(s_j), %
\end{equation}
\begin{equation}\label{eqn:theta}
\begin{split}
\phi(i,j) & = \mathtt{softmax}\{\mathcal{D}(s_i,s_j)\} \\
& = \frac{exp(-\beta \cdot \mathcal{D}(s_i,s_j))}{\sum_{k:(i,k) \in \mathcal{E} } exp(-\beta \cdot \mathcal{D}(s_i,s_k))  },
\end{split}
\end{equation}
where $ \beta = 0.1 $ is a smoothing factor. The spatially-pooled feature $ \Phi(s_i) $ is essentially a weighted average of the structural-appearance features ($ f_a $) of the 5 closest segments to segment $s_i$. This captures information on the immediate neighbourhood of $s_i$ and thus contributes towards encoding topological information in the scene.

\subsubsection{\textbf{Temporal Feature Pooling}}

Temporal relationships are defined by segment correspondences between frames. The segment $ s^{P_n}_i $ in the current point cloud frame $ P_n $ will only relate to its corresponding segment in each of the $ k_t $ previous frames ($ k_t = 3 $). Segment correspondence indices $ S_c $ are calculated as in Algorithm 1 iteratively $ k_t $ times. 

First $ kNN $ finds the indices of the k nearest neighbours of $ f_{a_i} $ from the set of features $ F^{P_{l-1}}_a $ of the previous frame. Next, $ rNN $ finds the indices of nearest neighbor centroids from the previous frame within a radius $ r = 1m $. 
To increase accuracy, the $ rNN $ search takes into account the the sensor's relative pose across frames represented by the homogeneous transformation $ \tensor[^{l-1}]{T}{_l} $. %
Finally, common elements in both nearest neighbour sets are found using the $ \mathtt{intersection} $ function and the function $ \mathtt{arg\_min} $ finds the index of the segment which minimises both feature-space and Euclidean-space distance.

\begin{algorithm}[t]
\SetAlgoLined
\KwResult{Corresponding segment indices $ S_c \in \mathbb{R}^m $}
 $ S^{P_l}, S^{P_{l-1}} \in \mathbb{R}^{m' \times n' \times 3} $ // segments\;
 $ F^{P_l}_a, F^{P_{l-1}}_a  \in \mathbb{R}^{m' \times d}$ // features\;
 $ C^{P_l}, C^{P_{l-1}} \in \mathbb{R}^{m' \times 4} $ // segment centroids\;
$ \tensor[^{l-1}]{T}{_l} \in SE(3) $ // relative pose from $ P_{l-1} $ to $ P_l $\;  
 \For{$ s_i \in S^{P_l} $}{
   $ f_{a_i}, \tensor[^{l}]{c}{_i}  $ // feature, centroid of segment $ s_i $ in $ P_l $\;
   $ F_{N} \leftarrow kNN(f_{a_i}, \ F^{P_{l-1}}_a ) $\;
   $ C_{N} \leftarrow rNN( \tensor[^{l-1}]{T}{_l}  \tensor[^{l}]{c}{_i}, \ C^{P_{l-1}} ) $\;
   $ {N} \leftarrow intersection(F_{N}, C_{N}) $\;
  \eIf{$ len(N) > 0$}{
   $ S_c[i] \leftarrow arg\_min(N) $ \;
   }{
   $ S_c[i] \leftarrow \varnothing $\;
  }
 }
 \caption{Estimate temporal segment correspondences from frame  $ P_{l} $ to $ P_{l-1} $}
\end{algorithm}

For the selected segment $ s^{P_n}_i $ in the current frame, the set of corresponding segments can be obtained from $ S_c $ as 
$  \{ s^{P_{n-1}}_i, .., s^{P_{n-k_c}}_i \} = \{ \mathbb{S}_{n,k_t}(j) : j \in S_c \}  $. We note that $ \ k_c \leq k_t  $ since correspondences can be lost between frames (when a segment which minimises both feature space and Euclidean-space distance is not available in the previous frame). The temporal pooling for $ s^{P_n}_i $ is then carried out as,
\begin{equation} %
\Psi(s_i) = \sum_{j \in S_c } \psi(i,j)f_{a}(\Tilde{s_j}) , \quad \Tilde{s_j} = \mathbb{S}_{n,k_t}(j),
\end{equation}
where $ \psi(i,j) $ is calculated similar to $ \phi(i,j) $ in Eq. \ref{eqn:theta} with $ \mathcal{D}(s_i,s_j) $ replaced by $ \|f_{a}(s_i) - f_{a}(\Tilde{s_j})\| $ and $ k $ is sampled from $ S_c $.
The aggregation of features across multiple sequential frames essentially magnifies the  weight of features corresponding to highly repeatable segments (segments which are extracted at every frame in the sequence and mapped to a similar point in the $ f_a $ feature space each time) and thus, inherently down-weights non-repeatable segments. 

The final sptiotemporal feature $ f_b $ is obtained as the average of a spatially-pooled feature $\Phi(s_i)$ and a temporally-pooled feature $\Psi(s_i)$,
\begin{equation}
f_b(s_i) = ( \Phi(s_i) + \Psi(s_i) ) / 2.
\end{equation}

\subsection{Second-order pooling}

Given a set of segments $ S = \{s_1, .., s_m\} $ and two complementary sets of features $ F_a \in \mathbb{R}^{m \times d} $ and $ F_b \in \mathbb{R}^{m \times d} $ the second-order pooling $ F^{O_2} $ of the features is defined as,
\begin{equation}\label{eqn:o2p}
 F^{O_2} = \{F^{O_2}_{xy}\} , \quad F^{O_2}_{xy} = \max_{s \in S } \ f_{xy}^{o_2}(s), 
\end{equation}
where $ F^{O_2} $ is a matrix with elements $ F^{O_2}_{xy} ( 1 \leq x,y \leq d) $ 
and $ f^{o_2}(s) = f_{a}(s)f_{b}(s)^T \in \mathbb{R}^{d \times d} $ is the outer product of the two complementary features of segment $ s $ ($ f_{a}(s), f_{b}(s) \in \mathbb{R}^{d} $). This accounts to taking the element-wise maximum of the second-order features of all segments in the scene. 

In order to make the scene descriptor matrix $ F^{O_2} $ more discriminative, it is decomposed in to singular values as $ F^{O_2} = U\lambda V $ and transformed non-linearly using the Power-Euclidean (PE) transform \cite{7439823}\cite{Li2017} into $ F^{O_2}_{\alpha} $ by raising each of its Eigen values by a power of $ \alpha $ as follows,
\begin{equation}\label{eqn:o2p_nlt}
 F^{O_2}_{\alpha} = U\hat{\lambda}V, \quad  \hat{\lambda} = \textit{diag}(\lambda^{\alpha}_{1,1},  .. , \lambda^{\alpha}_{d,d}),
\end{equation}
where $ \alpha = 0.5 $. The matrix $ F^{O_2}_{\alpha} $ is flattened and normalized to obtain the final global descriptor vector \textbf{g}  $ \in \mathbb{R}^{d^2} $.

The aggregation of multi-level features using higher-order pooling has multiple advantages for segment-based place description. First, it allows the encoding of complementary features for each segment thus enabling the incorporation of structural-appearance information along with topological and temporal information. Second, even though different point clouds will consist of varying number of segments $ m $ of varying sizes $ N_i $,  the output dimension of the aggregated feature is fixed and therefore computational time can be greatly reduced. Finally, the aggregated feature is invariant to the permutation of its inputs, which results in a viewpoint invariant global descriptor for place recognition. 

%% file: chapters/experiments.tex
\section{Experimental Setup}
\label{sec:experiments}

\subsection{Evaluation criteria}
We evaluate performance using the Precision-Recall curve and its scalar metric: the maximum $ F1 $ score ($ F1_{max} $). Additionally, we use the Extended Precision ($ EP $) metric proposed in \cite{Ferrarini2020} as it highlights the maximum recall at 100\% precision which is vital for robustness in place recognition. For Precision (P) and Recall (R), the $ F1_{max} $ and $ EP $ scores are defined as,
\begin{equation}
 F1_{max} = \max_{  \tau } \ 2\frac{P_{\tau} \cdot R_{\tau}}{P_{\tau} + R_{\tau}}, \quad   EP = \frac{P_{R0} + R_{P100}}{2},
\end{equation}\label{eq:f1_ep}
where  $ \tau $ is the threshold for positive prediction, $ P_{R0} $ is the Precision at minimum Recall, and $ R_{P100} $ is the maximum Recall at 100\% Precision.

Retrieval is performed based on the comparison of the cosine distance of the query descriptor with a database of descriptors of previously visited places. 
Inline with the evaluation criteria of \cite{Kong2020}, previous entries adjacent to the query by less than $ 30s $ time difference are excluded from search to avoid matching to the same instance.
The top-1 retrieval is considered a positive if the associated distance to query is less than the test threshold ($ \tau $). 
A positive retrieval is considered a true-positive if it is less than 3m from the ground truth pose of the query and a false-positive if it is greater than 20m away to maintain consistency with the evaluation in \cite{Kong2020}.

\subsection{Dataset}
Evaluation is performed on the KITTI odometry dataset \cite{Geiger2013} which consists of Velodyne  HDL-64E LiDAR scans collected from a moving vehicle in multiple dynamic urban environments in Karlsruhe, Germany. In line with the most recent evaluation on this dataset \cite{Kong2020}, we evaluate on sequences 00, 02,  05,  06,  07,  and  08. 

%% file: chapters/results.tex
\section{Results}
\label{sec:results}

In this section, we demonstrate the contribution of each component in our proposed method. We also provide quantitative and qualitative results to validate our method in comparison to state-of-the-art methods and we quantitatively evaluate the robustness our proposed method against viewpoint changes and sensor occlusion.

\begin{table}[t]
\centering
\begin{tabular}{c c c} 
 \hline
 Feature Types & $ EP $ & $ F1_{max} $  \\ [0.5ex] 
 \hline
  structural $ \otimes $ structural   & 0.739 & 0.796  \\
 structural $ \otimes $ spatial   & 0.847 & 0.895 \\
 structural $ \otimes $ temporal   & 0.865 & 0.900 \\
 \textbf{structural $ \otimes $ spatiotemporal}    & \textbf{0.912} & \textbf{0.931} \\
 \hline
\end{tabular}
\caption{A comparison of \textit{Locus} performance on KITTI sequence 08 when different input feature types are used. }
\label{table_ablation}
\end{table}

\subsection{Contribution of individual components }

We investigate the contributions of each component (ablation study) of our proposed method using $ F1_{max} $ and $ EP $ on KITTI sequence $ 08 $.  First, we demonstrate the importance of the Power Euclidean (PE) non-linear transform in our second-order pooling. For the generation of feature $ f_a $, we use the model trained on extracted segments from the sequences 05 and 06 \cite{Dube2020}.  The most basic second-order pooling method (\textit{O2P}) is the aggregation of second-order statistics of the same feature (Eq. \ref{eqn:o2p} with $f_b = f_a $). This can be extended with the Power Euclidean (PE) non-linear transform to get \textit{O2P+PE} (Eq. \ref{eqn:o2p_nlt} with $f_b = f_a $).
The \textit{O2P+PE} shows a dramatic increase of 44.7\% $F1_{max}$ (from 0.550 to 0.796) and and improvement of 32.4\% $ EP $ (from 0.564 to 0.739) when including the non-linear transform. This demonstrates the importance of PE for obtaining discriminative descriptors.

The rest of the evaluation uses the \textit{O2P+PE} feature pooling and compares the contribution of each input feature type. 
The input $ f_a $ is the structural appearance feature and the input feature $ f_b $ is varied across all multi-level feature types.  Table \ref{table_ablation} shows the respective performance for each of the following cases: $f_b = f_a $ (structural), $f_b = \Phi $ (structural $ \otimes $ spatial), $f_b = \Psi $ (structural $ \otimes $ temporal), and $f_b = ( \Phi + \Psi ) / 2 $ (structural $ \otimes $ spatiotemporal).

Table \ref{table_ablation}  demonstrates that pooling information from complementary features improves the place recognition performance. Furthermore, incorporating topological or temporal information individually makes the final descriptor more discriminative and improves $F1_{max}$ by 12.4\% and 13.1\% respectively. The best performance of our proposed method is achieved when incorporating both topological and temporal information along with the structural appearance (+ 17\% $F1_{max}$). 

\begin{figure}[t]
    \centering
    \includegraphics[width=0.3\textwidth]{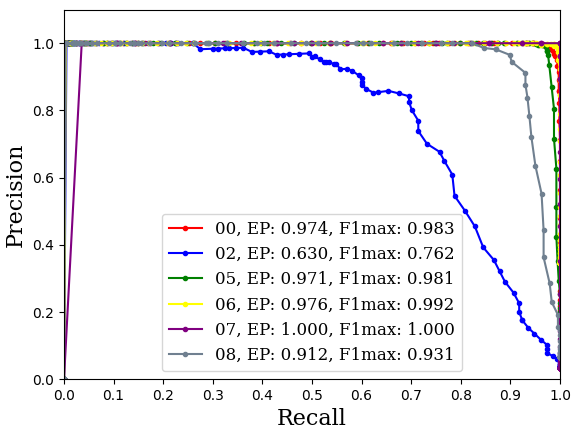}
    \caption{Precision-Recall curves of our proposed \textit{Locus} method on the KITTI dataset.}
    \label{fig:final_pr_curves}
\end{figure}

\begin{table*}[t]
\centering
\begin{tabular}{c | c c c c c c | c} 
 \hline
 Methods & 00 & 02 & 05 & 06 & 07 & 08 & Mean \\ [0.5ex] 
 \hline
  M2DP \cite{He2016} & 0.836 & 0.781 & 0.772 & 0.896 & 0.861 & 0.169 & 0.719 \\ 
 ScanContext \cite{Kim2018} & 0.937 & 0.858 & 0.955 & \textbf{0.998} & 0.922 & 0.811 & 0.914 \\ 
 PointNetVLAD \cite{Uy2018} & 0.785 & 0.710 & 0.775 & 0.903 & 0.448 & 0.142 & 0.627 \\ 
 PointNetVLAD* \cite{Uy2018} & 0.882 & 0.791 & 0.734 & 0.953 & 0.767 & 0.129 & 0.709 \\ 
 SemGraph-RN \cite{Kong2020} & 0.960 & \textbf{0.859} & 0.897 & 0.944 & 0.984 & 0.783 & 0.904 \\ 
  \textbf{\textit{Locus} (Ours)} & \textbf{0.983} & 0.762 & \textbf{0.981} & 0.992 & \textbf{1.0} & \textbf{0.931} & \textbf{0.942} \\ [0.5ex] 
 \hline
\end{tabular}
\caption{$ F1_{max} $ scores on the KITTI dataset. Results of M2DP, ScanContext, PointNetVLAD, and SemGraph are as presented in \cite{Kong2020} with the same evaluation criteria. PointNetVLAD* is the PointNetVLAD method retrained on KITTI \cite{Kong2020}.
}
\label{table_full_results}
\end{table*}

\begin{figure*}[h]
    \centering
    \includegraphics[width=\textwidth]{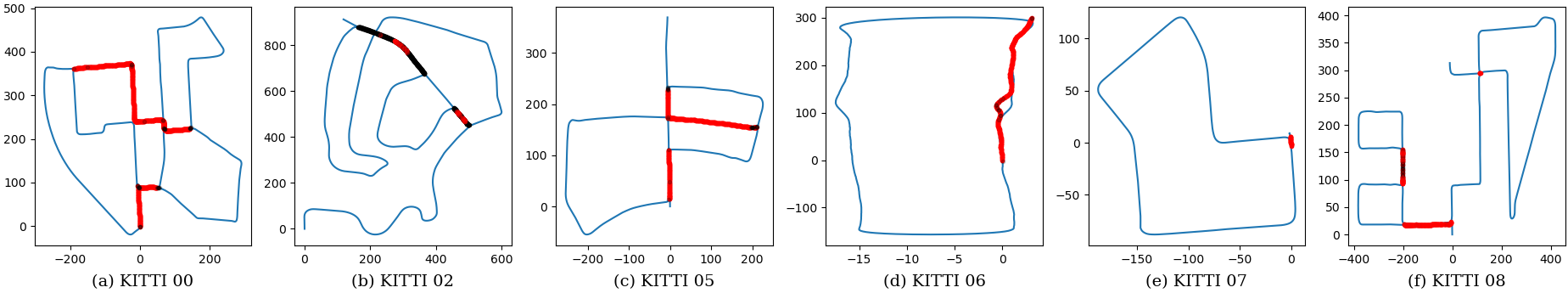}
    \caption{Qualitative performance visualization of our method at $ R_{P100} $ (zero false positives) along the trajectory. Red: true positives, Black: false negatives, Blue: true negatives (not revisits). }
    \label{fig:path_loc}
\end{figure*}

\subsection{Comparison to State-of-the-Art}

In this section, we compare the results of our method against other state-of-the-art results. 
Table \ref{table_full_results} shows summary comparison of $ F1_{max} $ scores. \textit{Locus} outperforms all other methods with the highest mean $ F1_{max} $ score (0.942). 
Our method sets a new state-of-the-art place recognition performance on KITTI dataset. %
\textit{Locus} achieves a significant improvement on the challenging sequence $ 08 $ which consists of number of reverse or orthogonal revisits thereby demonstrating the robustness to viewpoint variations. 

The Precision-Recall curves of \textit{Locus} across all sequences are presented in Fig. \ref{fig:final_pr_curves}.  The curves highlight that the performance of \textit{Locus} on sequence $ 02 $ is significantly poor compared to other sequences. As depicted in Fig. \ref{fig:path_loc}, sequence $ 02 $ has a long stretch of road with many false negatives. This road consists of non-descriptive point clouds where the segment extraction process fails to obtain a high number of descriptive segments. 

The qualitative visualization of our method at $ R_{P100} $ (\ie{} maximum Recall at 100\% Precision) is shown in Fig. \ref{fig:path_loc}. The figure shows the locations of true positive (red) and false negative (black) retrievals along the trajectory. This demonstrates that our method is capable of obtaining accurate retrievals in a wide range of environment and revisit types. 
We observe that most failure cases (black dots) occur near intersections. %

\subsection{Robustness tests}

We evaluate the robustness of our method against the state-of-the-art on a variety of challenging scenarios which simulate real-word adverse conditions. We simulate these adverse conditions by introducing a set of distortions to the input point cloud. 
Performance is compared based on the $ F1_{max} $ metric on the same KITTI sequences.

\subsubsection{\textbf{Viewpoint changes}}
We simulate viewpoint changes by rotating point clouds about the z-axis with a random angle. 
The change in mean $ F1_{max} $ over all sequences on this test compared to the normal results was -0.007 for our method. The respective change for ScanContext was +0.002 and for SemGraph-RN it was -0.003 as presented in \cite{Kong2020}. \textit{Locus}, ScanContext and SemGraph-RN show less than 1\% difference implying that random rotation of the input point cloud does not effect the final performance of these methods. 
Other methods such as M2DP, PointNetVLAD have been shown not to be rotation invariant as mentioned in \cite{Kong2020}. 

\subsubsection{\textbf{Occlusion}}
\begin{figure}[t]

    \centering
    \includegraphics[width=0.4\textwidth]{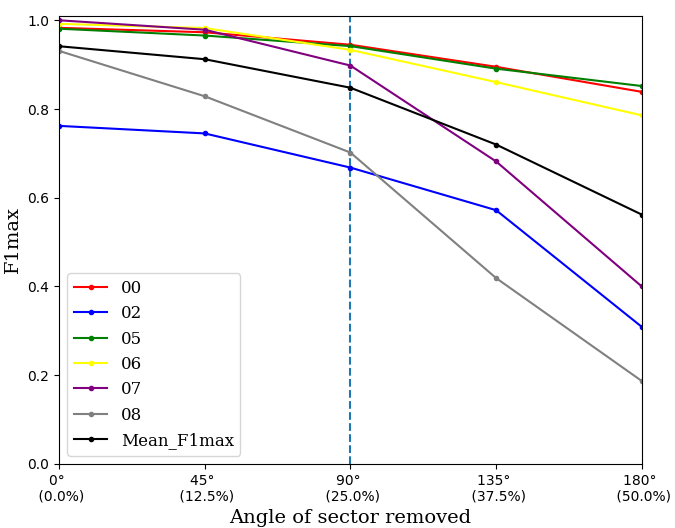}

\caption{Robustness of \textit{Locus} to scan occlusion. Vertical blue line shows when the mean $ F1_{max} $ drops by 10\%.}
\label{fig:rob_test_occ}
\end{figure}

This test aims to simulate occlusion of the LiDAR where the field-of-view of the sensor can be greatly reduced due to nearby dynamic objects or self-occlusion. 
We extend the occlusion test in \cite{Kong2020} which only considered occlusions of 30\textdegree \ by evaluating performance of various occlusion angles $ \theta_{occ} $  from 0\textdegree \ up to 180\textdegree. The occlusion test consists of removing all points which lie within a sector of a randomly selected azimuth. 
We compare our method with ScanContext (SC) \cite{Kim2018}, which achieved the second best overall performance in the Table~\ref{table_full_results}.  
ScanContext (SC) mean $F1_{max}$ performance drops by more than 20\% at 45\textdegree{} occlusion and around 50\% at 90\textdegree{} occlusion.  
Our method shows small performance degradation of 3.2\% at 45\textdegree{} occlusion and 9.98\% at 90\textdegree{} occlusion. Our method outperforms ScanContext (SC) with a large margin of 45.2\% mean $F1_{max}$ at  90\textdegree{} occlusion. 
Also note that in our method, the $ F1_{max} $ of sequences  $ 00 $ and $ 05 $ remains above 80\% even at occlusions of 180\textdegree{} where 50\% of the point cloud is removed. Results for all sequences are depicted in Figure \ref{fig:rob_test_occ}.

%% file: chapters/conclusion.tex
\section{Conclusion}
\label{sec:conclusion}

In this paper, we presented \textit{Locus}, a novel LiDAR-based place recognition method for large-scale environments.
We presented the advantages of scene representation via the aggregation of multi-level features related to components in a scene. We quantitatively showed how the inclusion of topological and temporal information in the description stage leads to an improvement in final place recognition performance. We formulated the generation of a global descriptor which incorporates all multi-level features without violating rotational-invariance. We validated our method through evaluation on the KITTI dataset where it surpassed the state-of-the art. Furthermore, we demonstrated the robustness of \textit{Locus} against viewpoint changes and occlusion.